\pdfoutput=1
\documentclass[sigconf,balance,nonacm]{acmart}
\usepackage{subfiles}
\usepackage{booktabs}
\usepackage{multirow}
\usepackage{stfloats}
\usepackage{lipsum}
\usepackage{arydshln}
\usepackage{graphicx}
\usepackage{subcaption}
\usepackage[normalem]{ulem}

\usepackage{amssymb}
\usepackage{tikz}
\usepackage{amsmath}
\usepackage{enumitem}
\usepackage{xcolor}

\definecolor{ao}{rgb}{0.0, 0.5, 0.0}
\newcommand{\vg}[1]{\textcolor{black}{#1}}
\newcommand{\vi}[1]{\textcolor{black}{#1}}
\newcommand{\vic}[1]{\textcolor{black}{#1}}
\newcommand{\zhiwei}[1]{\textcolor{black}{#1}}

\newcommand{\zhiweihu}[1]{\textcolor{black}{#1}}
\newcommand{\zwh}[1]{\textcolor{black}{#1}}

\newcommand{\jeff}[1]{\textcolor{black}{#1}}

\AtBeginDocument{%
  \providecommand\BibTeX{{%
    \normalfont B\kern-0.5em{\scshape i\kern-0.25em b}\kern-0.8em\TeX}}}

\setcopyright{acmcopyright}
\copyrightyear{2023}
\acmYear{2023}
\acmDOI{10.1145/3583780.3614922}

\acmConference[CIKM '23]{the 32nd ACM International Conference on Information and Knowledge Management}{October 21--25, 2023}{Birmingham, UK}

\acmBooktitle{Proceedings of the 32nd ACM International Conference on Information and Knowledge Management (CIKM '23), October 21--25, 2023, Birmingham, UK} 
\acmPrice{15.00}
\acmISBN{979-8-4007-0124-5/23/10}

\begin{document}

\title{HyperFormer: Enhancing Entity and Relation Interaction for Hyper-Relational Knowledge Graph Completion}

\author{Zhiwei Hu}
\affiliation{
  \institution{School of Computer and Information Technology \\ Shanxi University}
  \city{Taiyuan}
  \country{China}
}
\email{zhiweihu@whu.edu.cn}

\author{Víctor Gutiérrez-Basulto}
\affiliation{
  \institution{School of Computer Science and Informatics\\ Cardiff University}
  \city{Cardiff}
  \country{UK}
}
\email{gutierrezbasultov@cardiff.ac.uk}

\author{Zhiliang Xiang}
\affiliation{
  \institution{School of Computer Science and Informatics \\ Cardiff University}
  \city{Cardiff}
  \country{UK}
}
\email{xiangz6@cardiff.ac.uk}

\author{Ru Li}
\authornote{Contact Authors.}
\affiliation{
  \institution{School of Computer and Information Technology \\ Shanxi University}
  \city{Taiyuan}
  \country{China}
}
\email{liru@sxu.edu.cn}

\author{Jeff Z. Pan}
\authornotemark[1]
\affiliation{
  \institution{ILCC, School of Informatics\\ University of Edinburgh}
  \city{Edinburgh}
  \country{UK}
}
\email{j.z.pan@ed.ac.uk}



\renewcommand{\shortauthors}{Z. Hu, V Gutiérrez-Basulto, Z. Xiang, R. Li and J. Z. Pan}






\begin{abstract}
\vg{Hyper-relational knowledge graphs (HKGs) extend standard knowledge graphs by associating attribute-value qualifiers  to triples, which  effectively represent additional fine-grained information about its associated triple. Hyper-relational knowledge graph completion (HKGC) aims at inferring unknown triples while considering its qualifiers. Most existing approaches to HKGC exploit a global-level graph structure to encode hyper-relational knowledge into the graph convolution message passing process. However, the addition of multi-hop information might bring  noise into the triple prediction process. To address this problem,  we propose HyperFormer, a  model that considers  local-level sequential \zwh{information}, which  encodes the content of the entities, relations and qualifiers of a triple. More precisely, HyperFormer is composed of three different modules: an \textbf{\emph{entity neighbor aggregator}} module allowing to integrate the information of the     neighbors of an entity to capture different perspectives of it; a \textbf{\emph{relation qualifier aggregator}} module to integrate hyper-relational knowledge into the corresponding relation to refine the representation of relational content; a \textbf{\emph{convolution-based bidirectional interaction}} module based on a convolutional operation, capturing pairwise bidirectional interactions of entity-relation, entity-qualifier, and relation-qualifier.
Furthermore, we introduce a Mixture-of-Experts strategy into the feed-forward layers of HyperFormer to strengthen its representation capabilities while reducing the amount of model parameters and computation. Extensive experiments on three well-known datasets with four different conditions demonstrate HyperFormer's effectiveness.} Datasets and code are available at \url{https://github.com/zhiweihu1103/HKGC-HyperFormer}.

\end{abstract}

\maketitle

\section{Introduction}
\label{sec:intro}
\begin{figure*}[!htp]
    \centering
    \includegraphics[width=1\textwidth]{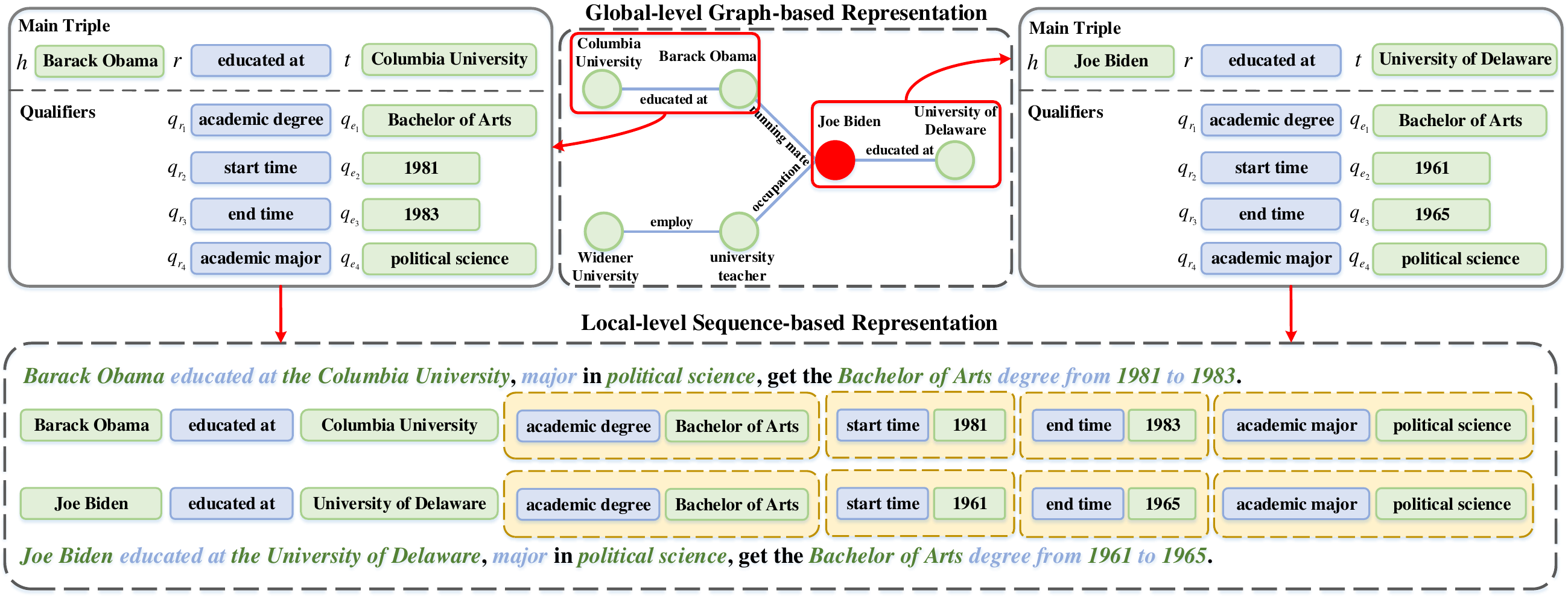}
    \caption{The global-level graph-based representation and local-level sequence-based representation based on two triples with qualifier pairs.}
    \label{figure_instane_and_mrr_degree}
\end{figure*}

\vic{Knowledge Graphs (KGs)~\cite{PVGW2017} store and organize factual knowledge of the world using triples of the form $(h,r,t)$~\cite{Pan2009}, capturing that  entities  $h,t$ are connected via  relation $r$. Popular KGs, such as WordNet~\citep{George_1995}, Freebase~\citep{Kurt_2008}, and Wikidata~\citep{Denny_2014}, are widely used in several tasks, ranging from question answering~\citep{FPKW2019,Apoorv_2020, Hongyu_2021, Zhiwei_2022} to recommendation systems~\citep{Xiang_2019, Tao_2021}. 
However,  \jeff{relational KGs} have no means for representing additional information of facts. For example, for the triple \zhiweihu{(\emph{Joe Biden}, \emph{educated at}, \emph{University of Delaware})} represented in Figure~\ref{figure_instane_and_mrr_degree}, it is non-trivial representing the major studied by Joe Biden at \zhiweihu{\emph{the University of Delaware}}. To address this shortcoming, \emph{hyper-relational KGs} have been proposed, extending \zhiweihu{ binary} \jeff{relational} KGs by  associating with each triple additional attributes in the form of   relation-entity
pairs, known as \emph{qualifiers}. \jeff{Thus} in this case a relational fact is composed by the main triple and its qualifiers. For example, for the triple in Figure~\ref{figure_instane_and_mrr_degree}, the qualifier pairs \zhiweihu{(\emph{academic major}, \emph{political science}), (\emph{academic degree}, \emph{Bachelor of Arts}), (\emph{start time}, \emph{1961}), and (\emph{end time}, \emph{1965})} describe the major and degree information of \emph{Joe Biden} education at \zhiweihu{\emph{the University of Delaware} from \emph{1961} to \emph{1965}}. Like standard KGs, hyper-relational KGs are inevitably incomplete. To tackle this problem, several hyper-relational knowledge graph completion (HKGC) approaches have been recently proposed~\citep{Paolo_2020,Mikhail_2020, Quan_2021, Harry_2022},} to examine the impact of the addition of qualifier pairs on the knowledge graph completion task.

\vg{Most existing methods for HKGC~\citep{Mikhail_2020, Quan_2021, Harry_2022} employ graph convolutional networks (GCNs) to incorporate qualifier pairs information into entity and relation embeddings. 
In particular, when encoding the content of the graph structure, these approaches  use multiple layers of graph convolution operations to incorporate multi-hop information into the representation of entities. Although these methods enrich the representation of entities, they inevitably introduce additional noise by considering information that might not be relevant for an entity. 
More precisely, the first source of noise comes from entities in the standard KG, i.e., entities occurring in main triples. For instance, in 
Figure~\ref{figure_instane_and_mrr_degree} in the \emph{Global-level Graph-based Representation} part, when trying to predict (\emph{Joe Biden}, \emph{educated at}, \emph{?}) using two layers of graph convolution operations, information about the enities \emph{Columbia University}, \emph{Barack Obama}, \emph{Widener University}, \emph{university teacher} will be incorporated into the representation of \emph{Joe Biden}. However, the confidence of the true answer \emph{University of Delaware}  will be affected by  the information from the entities \emph{Columbia University} and \emph{Widener University}, as the three schools share a high degree of similarity.
A second source of noise comes from the introduction of hyper-relational knowledge. 
Going back to the previous example, to predict the triple (\emph{Joe Biden}, \emph{educated at}, \emph{?}) with qualifiers (\emph{start time}, \emph{1961}), and (\emph{end time}, \emph{1965}), the  neighbor (\emph{Barack Obama}, \emph{educated at}, \emph{Columbia University}) includes the qualifiers (\emph{start time}, \emph{1981}) and (\emph{end time}, \emph{1983}), which will affect the representation of the relation \emph{educated at} and the prediction of where \emph{Joe Biden} was educated at.} 

\vg{The main objective of this paper is to introduce an alternative  to global graph operations for the HKGC task. Note that, as shown in Figure~\ref{figure_instane_and_mrr_degree}  in the \emph{Local-level Sequence-based Representation} part, the local sequential  content does not introduce redundant entity and relation content present in the global graph structure, because the entities and relations involved in a local sequence are directly related to the content to be predicted. 
With this in mind, we introduce \textbf{HyperFormer}, a  framework for HKGC that considers local-level sequential information and abandons the global-level structural content. Specifically HyperFormer integrates hyper-relational information into the entity and relation embeddings of a fact  by using  three modules: an \textbf{\emph{entity neighbor aggregator}} module allowing to integrate the information of  one-hop local neighbors of an entity into its representation to capture different perspectives of it;
a \textbf{\emph{relation qualifier aggregator}} module to integrate hyper-relational knowledge into the corresponding relation representation, so that relations occurring in different facts are contextualized by the qualifier pairs information; 
a \textbf{\emph{convolution-based bidirectional interaction}} module based on a convolutional operation, capturing pairwise bidirectional interactions of entity-relation, entity-qualifier, and relation-qualifier. Furthermore, to increase HyperFormer's capacity while reducing the amount of parameters and calculations, we introduce a \emph{Mixture-of-Experts (MoE)} strategy to leverage the sparse activation nature in the feed-forward layers of transformers. Our contributions can be summarized as follows:}

\begin{itemize}[itemsep=0.2ex, leftmargin=5mm]
\item We propose a framework for HKGC that fully exploits local-level sequential information, while preserving the structural information of qualifiers. Further, we integrate the information of one-hop neighbors of an entity to capture different perspectives of it. In addition, the adoption of a bidirectional interaction mechanism strengthens the awareness between entities, relations, and qualifiers.

\item We introduce a MoE strategy to enhance the  representation capabilities of HyperFormer, while reducing the number of parameters and calculations of the model.

\item 
We conduct extensive experiments with four different conditions: mixed-percentage mixed-qualifier, fixed-percentage mixed-qualifier, fixed-percentage fixed-qualifier, and different numbers of entity's neighbors. 
Our results show that HyperFormer achieves SoTA performance for HKGC. We also conducted various  ablation studies.
\end{itemize}

\section{Related Work}
\label{sec:relwork}
\subsubsection*{Knowledge Graph Completion.}
\vic{ 
There are mainly two kinds of existing KGC methods: structure-based methods and description-based methods. Depending on the type of embedding space, structure-based methods can be divided into three categories: (i) point-wise space methods, e.g., TransE~\citep{Antoine_2013}, TransR~\citep{Yankai_2015}, HAKE~\citep{Zhanqiu_2020}; (ii) complex vector space methods, e.g.\ ComplEx~\citep{TrouillonWRGB_2016}, RotatE~\citep{Zhiqing_2019}, QuatE~\citep{Shuai_2019}; (iii) manifold space methods, e.g., MuRP~\citep{Ivana_2019}, AttH~\citep{Ines_2020}. Description-based methods leverage text descriptions of entities and relations, e.g., KG-BERT~\citep{Liang_2019}, StAR~\citep{Bo_2021}, CoLE~\citep{Yang_2022}. 
There are also KGC approaches~\cite{Wiharja2020,geng2021} considering schema information. 
}
%

\subsubsection*{Hyper-relational Knowledge Graph Completion}
\vg{Earlier works on HKGC proposed embedding-based methods to  learn and reason with hyper-relational knowledge~\citep{Jianfeng_2016,Richong_2018,Saiping_2019,Yu_2021}. However, they assume that  hyper-relational knowledge is  equally important for all relations, which is often not the case. To encode the contribution of different qualifier pairs, HINGE~\citep{Paolo_2020} adopts a convolutional framework to iteratively convolve every  qualifier pair into the main triple, which can naturally discriminate the importance of different hyper-relational facts. StarE~\citep{Mikhail_2020} extends CompGCN~\citep{Shikhar_2020} by encoding the qualifier pairs of a triple and further combining  it with the  relation representation, and then uses a transformer~\citep{Ashish_2017} encoder to model the interaction between qualifiers and the main triple.  Hy-Transformer~\citep{Donghan_2021} replaces the computation-heavy GCN aggregation module with a layer normalization operation~\citep{Lei_2016}, significantly improving the computational efficiency. 
QUAD~\citep{Harry_2022}, based on the StarE model, proposes a framework that utilizes multiple aggregators to learn better representations for hyper-relational facts.
However, the global-level graph structure adopted by both StarE and QUAD integrates the multi-hop neighbor content into the corresponding entity through the graph convolution process, which inevitably introduces  noise, because the node content far away from an entity will affect the real representation of such entity.
HAHE~\citep{Haoran_2023} introduces the global-level and local-level attention to model the graphical structure and sequential structure, however, the introduction of graph structure information brings a large burden to model computation.
GRAN~\citep{Quan_2021} represents  hyper-relational facts as a heterogeneous graph, representing it with edge-aware attentive biases to capture both local and global dependencies within the given facts. In particular, GRAN  also considers a local sequential  representation structure and captures the   semantic information inside hyper-relation facts by using a transformer encoder~\citep{Ashish_2017}. However, GRAN has three shortcomings. First, it only considers the knowledge directly related to the current statement, fully ignoring any type of information from the neighbors of  an entity. 
Second, the constraining process of the hyper-relational knowledge onto the main triple is simply handed over to a transformer, without capturing the pair-like structure of qualifiers. 
Third, the transformer uses  full connection attention to realize the interaction between each token in the sequence, ignoring the \zwh{explicit} interaction between entities, relations and hyper-relational knowledge. }

\section{Method}
\vic{In this section, we describe the architecture of  HyperFormer (cf.~Figure~\ref{figure_model_structure}). We start by introducing necessary background (\S~\ref{section_background}), and then present in detail 
\jeff{its} modules. 
(\S~\ref{section_model_architecture}).}

\subsection{Background}
\label{section_background}
\subsubsection*{Hyper-relational Knowledge Graph.}
Let $\mathcal{V}$ and $\mathcal{R}$  be finite  sets of \emph{entities} and \emph{relation types}, respectively. Furthermore, let $\mathcal Q = 2^{(\mathcal R \times \mathcal V)}$. A \emph{hyper-relational knowledge graph} $\mathcal{G}$ is a tuple $(\mathcal V, \mathcal R, \mathcal T)$, where $\mathcal T$ is a finite set of \emph{(qualified) relational facts}. The relational facts in $\mathcal T$ are of the form $(h,r,t,qp)$ where $h, t \in \mathcal V$ are  the \emph{head} and \emph{tail} entities,  $r \in \mathcal R$ is the 
\jeff{\emph{relation}} connecting  
\jeff{$h$ and $t$},
and $qp =\{(q_{r_1},q_{e_1}), \ldots, (q_{r_n},q_{e_n})\}  \in \mathcal Q$ is a set of \emph{qualifier pairs}, with \emph{qualifier relations} $q_{r_i} \in \mathcal R$   and \emph{qualifier entities $q_{e_i} \in \mathcal V$}. We will refer to $(h,r,t)$ as the \emph{main triple} of the relational fact. Under this representation regime, we can enrich the main triple (\emph{Joe Biden}, \emph{educated at}, \zhiweihu{\emph{University of Delaware}}) in Figure~\ref{figure_instane_and_mrr_degree}  with the additional semantic information provided by the qualifiers as follows: (\emph{Joe Biden}, \emph{educated at}, \emph{University of Delaware}, (\emph{academic degree}, \emph{Bachelor of Arts}), (\emph{academic major}, \emph{political science}), (\emph{start time}, \emph{1961}), (\emph{end time}, \emph{1965})). Crucial to our approach is the information provided by the neighbors of an entity:  For an entity $h$, we define its \textit{neighbors} $\mathcal{N}_h = \{(r,t,qp) \mid (h,r,t,qp) \in \mathcal{G}\}$.

\label{section_background_hkg}


\subsubsection*{Hyper-relational Knowledge Graph Completion.}
\vg{Following previous work,   similar to the KG completion task,  the HKGC task  aims at  predicting the correct head or tail entity in a fact. More precisely, given a relational fact $(h,r,?,qp)$ or $(?,r,t,qp)$ with the tail or head entity of the main triple missing, the aim is to infer the missing entity $?$ from $\mathcal V$.}

\subsection{Model Architecture}
\label{section_model_architecture}

\vic{
We present in this section the three modules of HyperFormer: $(i)$~\emph{Entity Neighbor Aggregator}, \S \ref{section_ena}; $(ii)$ \emph{Relation Qualifier Aggregator}, \S \ref{section_rqa}; $(iii)$ \emph{Convolution-based Bidirectional Interaction}, \S \ref{section_cbi}. Furthermore, we introduce a Mixture-of-Experts  strategy,  \S \ref{section_moe}.}

\subsubsection{ENA: Entity Neighbor Aggregator}
\label{section_ena}

\vic{We explore a transformer mechanism (previously used for other KG-reated tasks~\citep{Sanxing_2021, Yang_2022, Zhiwei_2022_2}) for the  HKGC task.   Given a relational fact $(h,r,t,qp)$, for  predicting the tail entity $t$, similar to the input representation of BERT~\citep{Jacob_2019}, we build an input sequence $S =(h,r, \texttt{[MASK]}, qp)$, where \texttt{[MASK]} is a special token  used in place of entity $t$. We randomly initialize each token input vector to feed it into a transformer and get the output representation $\textbf{\textit{E}}^{[\emph{mask}]}$ of the  \texttt{[MASK]} token:}
\begin{equation}
(\textbf{\textit{E}}^{h},\textbf{\textit{E}}^{r},\textbf{\textit{E}}^{[\emph{mask}]},\textbf{\textit{E}}^{qp}) = \texttt{Trm}(h,r, \texttt{[MASK]},qp)
\end{equation}
\vic{The representation $\textbf{\textit{E}}^{[\emph{mask}]}$  captures the information interaction between $h,r$ and $qp$. We will then use $\textbf{\textit{E}}^{[\emph{mask}]}$  to score all candidate entities and infer the most likely tail entities. }

\begin{figure*}[!htp]
    \centering
    \includegraphics[width=1\textwidth]{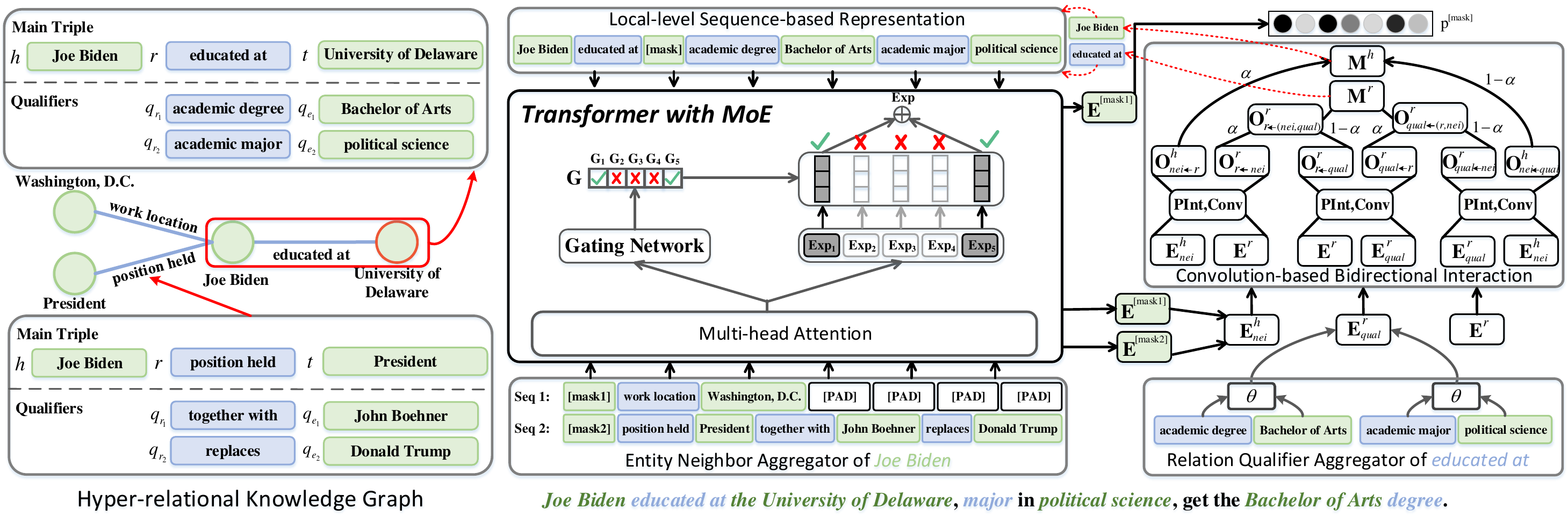}
    \caption{An overview of our HyperFormer model, containing three  modules:  Entity Neighbor Aggregator  (\S \ \ref{section_ena}), Relation Qualifier Aggregator  (\S \ \ref{section_rqa}), and Convolution-based Bidirectional Interaction  (\S \ \ref{section_cbi}).}
    \label{figure_model_structure}
\end{figure*}

\paragraph{\vic{Enhanced  Representation of Entity Neighbors.}} 
\vic{The one-hop neighbors of an entity help to describe  it from different perspectives. In many cases, simultaneously considering the information from multiple neighbors of an entity is necessary to correctly infer the entities that are related to it. For instance,  in Figure~\ref{figure_model_structure}, using  the  connections (\emph{position held}, \emph{President}) and (\emph{work location}, \emph{Washington, D.C.}) from \emph{Joe Biden}, we can infer that \emph{Joe Biden} is the president of \emph{the United States}, but not the prime minister of \emph{United Kingdom}. 
\vg{To adequately encode the content of the  neighbors of an entity, we introduce the \textbf{E}ntity \textbf{N}eighbor \textbf{A}ggregator (\textbf{ENA}) module. Given  a masked  tuple $(h,r, [\texttt{MASK}], qp)$, besides considering the embeddings of $h, r$ and $qp$, we also introduce information about the neighbors of $h$ as follows:}}
%

\vic{\begin{enumerate}[itemsep=0.5ex, leftmargin=5mm]
\item For all relational facts in $\mathcal G$ having $h$ as the head, we generate masked 4-tuples by using the placeholder \texttt{[MASK]} in place of $h$. More precisely, fix  entity $h$, we define the following set of neighbors of $h$: $\mathcal N_h = \{(\texttt{[MASK]},r',t',qp') \mid (h,r',t',qp') \in \mathcal G\}$. Note that for different tuples of the form $(h,r',t',qp')$ with $h$ as the head, different \texttt{[MASK]} representations will be obtained. For each \zhiweihu{tuple} $N \in \mathcal N_h$, the masked token representation of $N$ is denoted as $\textbf{\textit{N}}_h^{[\textit{mask}]}$.
%
%
\item We then sum up all \texttt{[MASK]} representations of the elements in $\mathcal N_h$ and further average the results to obtain the aggregated representation of the neighbors of $h$ as $\zhiwei{\textbf{\textit{E}}^{\, h}_{\textit{nei}}}=\textit{mean}(\sum\limits_{  N \in \mathcal{N}_h}{{\! \textbf{\textit{N}}}^{[\textit{mask}]}_h})$.
\end{enumerate}}

\subsubsection{RQA: Relation Qualifier Aggregator}
\label{section_rqa}
\vic{As discussed, qualifiers allow to describe relational content in a fine-grained manner. For example, in Figure~\ref{figure_model_structure} the qualifier pairs (\emph{academic major}, \zhiweihu{\emph{political science})} and (\emph{academic degree}, \zhiweihu{\emph{Bachelor of Arts}}) can respectively describe the major and degree information of the relation \emph{educated at} in the main triple (\emph{Joe Biden}, \emph{educated at}, \zhiweihu{\emph{University of Delaware}}). \zhiweihu{Directly serializing the qualifier content $qp$ into the main triple $(h,r, \texttt{[MASK]})$ can form a tuple representation $(h,r, \texttt{[MASK]}, qp)$, but this representation destroys the structural information of the qualifiers content. For example, \emph{political science} can limit the major that \emph{Joe Biden} obtained at the \emph{University of Delaware} when it appears at the same time as \emph{academic major}. }
To incorporate the sequence $(h,r, \texttt{[MASK]}, qp)$ along with the hyper-relational knowledge into the message passing process \zhiweihu{without damaging the qualifier's structural knowledge}, we introduce the \textbf{R}elation \textbf{Q}ualifier \textbf{A}ggregator (\textbf{RQA}) module. \zhiweihu{Specifically}, we use the following three steps to obtain the aggregated  representation of qualifier pairs for the  relation occurring in a  main triple:}
\vic{\begin{enumerate}[itemsep=0.5ex, leftmargin=5mm]
\item For a  relational fact  $(h,r,\zhiweihu{\texttt{[MASK]}},qp)$ with $qp =\{(q_{r_1}, q_{e_1}), \ldots,$ $(q_{r_n}, q_{e_n})\}$, 
we randomly initialize the embedding of \zhiweihu{qualifier relations and qualifier entities} for each qualifier pair, getting the input embedding: $\zhiwei{\textbf{\textit{qp}} = \{(\textbf{ \textit{q}}_{\textbf{\textit{r}}_1},\textbf{\textit{q}}_{\textbf{\textit{e}}_1}), \ldots, \textbf{ (\textit{q}}_{\textbf{\textit{r}}_n},\textbf{\textit{q}}_{\textbf{\textit{e}}_n})\}}$.
\item
Qualifier pairs  provide additional complementary relational knowledge, each of them  capturing different aspects of  it.
So, we aim to acquire a representation of $r$ given the knowledge of a qualifier pair $\textit{q}_{\textit{r}_i}$ and $\textit{q}_{\textit{e}_i}$.
To this end, we  consider both  a relation $r$  and a qualifier pair $(\textit{q}_{\textit{r}_i}, \textit{q}_{\textit{e}_i})$ as a form of pseudo-triple $(r, \textit{q}_{\textit{r}_i}, \textit{q}_{\textit{e}_i})$. 
We can then use several knowledge representation functions based on embedding methods to get a representation of $r$:  We compose the representations of the qualifier relations $\zhiwei{\textbf{\textit{q}}_{\textbf{\textit{r}}_i}}$ and qualifier entities $\zhiwei{\textbf{\textit{q}}_{\textbf{\textit{e}}_i}}$ using an entity-relation function $\theta$, such as TransE~\citep{Antoine_2013}, DistMult~\citep{Bishan_2015}, ComplEx~\citep{TrouillonWRGB_2016} or  RotatE~\citep{Zhiqing_2019}. We denote this composition as $\zhiwei{\Theta_i=\theta(\textbf{\textit{q}}_{\textbf{\textit{r}}_i}, \textbf{\textit{q}}_{\textbf{\textit{e}}_i})}$.  
\item The representations of different qualifier pairs are aggregated via a position-invariant summation function, which is then averaged: $\zhiwei{\textbf{\textit E}^{r}_{\textit{qual}}}=\textit{mean}(\sum_{i=1}^n{\Theta_i})$.
\end{enumerate}}
\subsubsection{CBI: Convolution-based Bidirectional Interaction} 
\label{section_cbi}
\vic{To obtain enhanced entity and relation representations, we could directly combine $\textbf{\textit{E}}^{\emph{h}}$ and $\zhiwei{\textbf{\textit{E}}^{h}_{\textit{nei}}}$, $\textbf{\textit{E}}^{r}$ and $\zhiwei{\textbf{\textit{E}}^{r}_{\textit{qual}}}$ using the addition operation. However, this simple fusion method cannot fully realize the deep interaction between entity, relation and qualifier pairs. Indeed, it is not enough to simply pass the message between the three of them to the transformer, \zhiweihu{because its fully-connected attention layer only captures universal inter-token associations}. To address this shortcoming, we propose a novel \textbf{C}onvolution-based \textbf{B}idirectional \textbf{I}nteraction (\textbf{CBI}) module to explicitly integrate each state of pairwise representation pairs: \emph{entity-relation}, \emph{entity-qualifiers} and \emph{relation-qualifiers}. For instance, to obtain the new relational embedding, the information of the entity neighbor embedding $\zhiwei{\textbf{\textit{E}}^{h}_{\textit{nei}}}$ and qualifier embeddings $\zhiwei{\textbf{\textit{E}}^{r}_{\textit{qual}}}$  can be integrated into the relation embedding $\textbf{\textit{E}}^{r}$ using the following four steps: }

\begin{enumerate}[itemsep=0.5ex, leftmargin=5mm]
\item \vg{We combine $\textbf{\textit{E}}^{r}$ and $\zhiwei{\textbf{\textit{E}}^{\textit{h}}_{\textit{nei}}}$ based on the convolution operation $\textbf{Conv}(\circ,\circ)$, and then pass the obtained joint representation to a  \emph{perceptron interaction layer} $\textbf{PInt}(\circ)$ 
to obtain $\textbf{\textit{O}}^r_{r\leftarrow\textit{nei}}$ and $\textbf{\textit{O}}_{\textit{nei}\leftarrow r}^h$ as follows:}
\begin{equation}
[\textbf{\textit{O}}^{r}_{r\leftarrow\textit{nei}};\textbf{\textit{O}}_{\textit{nei}\leftarrow r}^h]=\textbf{PInt}(\textbf{Conv}(\textbf{\textit{E}}^{r},\zhiwei{\textbf{\textit{E}}^{\textit{h}}_{\textit{nei}}}))
\end{equation}
\vg{We analogously  fusion $\textbf{\textit{E}}^{r}$ with $\zhiwei{\textbf{\textit{E}}^{r}_{\textit{qual}}}$ and $\textbf{\textit{E}}^{h}_{\textit{nei}}$ with $\textbf{\textit{E}}^{r}_{\textit{qual}}$ to respectively get $\textbf{\textit{O}}^{r}_{r\leftarrow\textit{qual}}$ and $\textbf{\textit{O}}_{\textit{qual}\leftarrow r}^r$, 
$\textbf{\textit{O}}^{h}_{\textit{nei}\leftarrow\textit{qual}}$ and $\textbf{\textit{O}}_{\textit{qual}\leftarrow \textit{nei}}^r$:}
\begin{equation}
[\textbf{\textit{O}}^{r}_{r\leftarrow\textit{qual}};\textbf{\textit{O}}_{\textit{qual}\leftarrow r}^r] =\textbf{PInt}(\textbf{Conv}(\textbf{\textit{E}}^{r},\zhiwei{\textbf{\textit{E}}^{r}_{\textit{qual}}}))
\end{equation}
\begin{equation}
[\textbf{\textit{O}}^{h}_{\textit{nei}\leftarrow\textit{qual}};\textbf{\textit{O}}_{\textit{qual}\leftarrow \textit{nei}}^r]=\textbf{PInt}(\textbf{Conv}(\textbf{\textit{E}}^{h}_{\textit{nei}},\textbf{\textit{E}}^{r}_{\textit{qual}}))
\end{equation}
\vic{The used $\textbf{Conv}(\circ,\circ)$ operation, for the initial fusion of two vectors, is \zhiweihu{similar to the one proposed by InteractE~\citep{Shikhar_2020_2}, but in principle  any other vector technique could be used instead}. We use a one-layer MLP as our $\textbf{PInt}(\circ)$ operation. Note that the result of $\textbf{PInt}$ is divided into two parts to obtain two enhanced vector representations. For example, for the integration of $\textbf{\textit{E}}^{r}$ and $\textbf{\textit{E}}_{\textit{qual}}^r$, after the two above operations are performed, we obtain the qualifier-aware relation representation $\textbf{\textit{O}}^{r}_{r\leftarrow\textit{qual}}$ and relation-aware qualifier representation $\textbf{\textit{O}}_{\textit{qual}\leftarrow r}^r$. 
These representations are defined in a bidirectional way, \zhiweihu{so} each of them contributes to the definition of the other.}
\item \vic{We then employ a  gating mechanism to combine both the  \zhiweihu{entity's} neighbor-aware relation representation $\textbf{\textit{O}}^{r}_{r\leftarrow\textit{nei}}$ and the qualifier-aware relation representation  $\textbf{\textit{O}}^{r}_{r\leftarrow\textit{qual}}$. The final representation of relation $r$ is \zhiweihu{denoted} as $\textbf{\textit{O}}^{r}_{r\leftarrow\textit{(nei,qual)}}$: 
\begin{equation}
\textbf{\textit{O}}^{r}_{r\leftarrow\textit{(nei,qual)}}=\alpha \odot \textbf{\textit{O}}^{r}_{r\leftarrow\textit{nei}} + (1-\alpha)\odot\textbf{\textit{O}}^{r}_{r\leftarrow\textit{qual}}
\end{equation}
\begin{equation}
\alpha=\sigma(W_1\textbf{\textit{O}}^{r}_{r\leftarrow\textit{nei}}+ W_2\textbf{\textit{O}}^{r}_{r\leftarrow\textit{qual}}+ b_1+ b_2)
\end{equation}
where $\alpha$ is the reset gate that controls the flow of information from $\textbf{\textit{O}}^{r}_{r\leftarrow\textit{nei}}$ to $\textbf{\textit{O}}^{r}_{r\leftarrow\textit{qual}}$, \zhiwei{$\sigma$ is the sigmoid function} and  ${W_1, W_2, b_1,b_2}$ are the parameters to \zhiweihu{be learned}.
}
%
%

\item 
\zhiwei{Similarly,} we get the relation-aware and qualifier-aware \zhiweihu{entity's neighbor} representation $\textbf{\textit{O}}^{h}_{\textit{nei}\leftarrow(r,\textit{qual)}}$, and  the relation-aware and \zhiweihu{entity's neighbor-aware} qualifier representation $\textbf{\textit{O}}^{r}_{\textit{qual}\leftarrow(r,\textit{nei)}}$. \zhiweihu{Then we add up  $\textbf{\textit{O}}^{r}_{r\leftarrow\textit{(nei,qual)}}$ and $\textbf{\textit{O}}^{r}_{\textit{qual}\leftarrow(r,\textit{nei)}}$ to obtain the final relational  representation $\textbf{\textit{M}}^{r}$. We use $\textbf{\textit{O}}^{h}_{\textit{nei}\leftarrow(r,\textit{qual)}}$ as the final entity representation, denoted as $\textbf{\textit{M}}^{h}$, which is the result of combining $\textbf{\textit{O}}^{h}_{nei\leftarrow\textit{r}}$ and $\textbf{\textit{O}}^{h}_{nei\leftarrow\textit{qual}}$ using Step (2).}
%

\item 
For the input masked fact $(h,r,  \texttt{[MASK]}, qp)$, after performing the above three steps, we get  enhanced representations  $\textbf{\textit{M}}^{h}, \textbf{\textit{M}}^{r}$ of the entity $h$ and relation $r$, which can respectively be used as the input initialization (together with  the randomly initialized \texttt{[MASK]} and $qp$) to the \zhiweihu{transformer encoder}. The output $\textbf{\textit{E}}^{[\emph{mask}]}$ is then used to score all candidate entities. \vic{More precisely, we use a standard softmax classification layer to predict the target entity, and use cross-entropy between the one-hot label and the prediction as training loss, defined as:}


\begin{equation}
p^{[\emph{mask}]}=\textit{softmax}(\textbf{\textit{E}}^{\textit{ent}}\textbf{\textit{E}}^{[\emph{mask}]})
\end{equation}
\begin{equation}
\mathcal{L}=-\sum_{mask}{y^{[\emph{mask}]}{{\rm log}\,p^{[\emph{mask}]}}}
\end{equation}

$\textbf{\textit{E}}^{\emph{ent}}$ represents  the embedding matrix of all entities in the dataset, \zhiwei{$p^{[\emph{mask}]}$} is the predicted distribution of the \texttt{[MASK]} position over all entities, $y^{[\emph{mask}]}$ is the corresponding one-hot label of the \texttt{[MASK]} position. 
\end{enumerate}

\subsubsection{Transformer with Mixture-of-Experts}
\label{section_moe}
\vic{Although transformers can achieve good results in many fields,  a recognized challenge for them is that the model parameters grow quadratically as the embedding dimension increases. However, it has been noted that two-thirds of the  parameters of a transformer are concentrated in the feed-forward layers (FFN), and that not all of them are necessary~\citep{Xiao_2022}.
%
To limit the training burden while increasing the model size, we introduce a Mixture-of-Experts (MoE)~\citep{Noam_2017} strategy into the transformer, which will help selecting the necessary parameters through a gating mechanism. More precisely, given an input $x$, MoE includes $n$ expert networks}
\vic{
$\{\textbf{Exp}_1(x),\textbf{Exp}_2(x),...,\textbf{Exp}_n(x)\},$
with
$\textbf{Exp}_i(x)\in \mathbb{R}^{d\times d}$, for all $ 1 \leq i \leq n$}
\vic{
and a gating network $\textbf{G}(x)\in \mathbb{R}^{d\times n}$ used to select specific experts: 
the \emph{i-th} element $\textbf{G}_i(x)$ in $\textbf{G}(x)$ specifies whether the  expert $\textbf{Exp}_i(x)$ should be selected. The output of the MoE module is calculated as follows:}
\begin{equation}
\begin{array}{rcl}
\textbf{Exp}=\sum_{i=1}^{n}\textbf{G}_i(x)\cdot\textbf{Exp}_i(x)\\[1mm]
\textbf{G}(x)={\textit{softmax}}(x\cdot \textbf{\textit{W}}_{\textit{gate}})\\[1mm]
\textbf{Exp}_i(x)=x(\textbf{\textit{W}}_i + b_i)\textbf{\textit{W}}_i^\prime + b_i^\prime\\
\end{array}
\end{equation}
\vic{ $\textbf{\textit{W}}_i, \textbf{\textit{W}}_i^\prime  \in \mathbb{R}^{d\times d}$ are the learnable parameters, $\textbf{\textit{W}}_{\textit{\!gate}}\in \mathbb{R}^{d\times n}$ is a trainable matrix with \emph{n}  the number of experts, and each expert $\textbf{Exp}_i(x)$ corresponds to a FFN. $\textbf{G}_i(x)\in \mathbb{R}^{d\times 1}$ is the value of \textbf{G} at the \emph{i-th} position on the \emph{2nd} dimension. In practice, we set it to 0 or 1 depending on whether the value of $\textbf{G}_i(x)$ exceeds a certain threshold, so $\textbf{G}(x)$ is sparse. We may set a large number of experts, but for each sample only $k$ of them are selected, called top experts. In the experimental part, we set \emph{n=64} and \emph{k=2}.}

\section{Experiments}
\begin{table}[!htp]
\setlength{\abovecaptionskip}{0.03cm} 
\renewcommand\arraystretch{1.2}
\setlength{\tabcolsep}{0.65em}
\centering
\small
\caption{Statistics of datasets \zhiweihu{under mixed-percentage mixed-qualifier and fixed-percentage mixed-qualifier scenarios}. The values in parentheses indicate that the corresponding percentage in corresponding dataset has hyper-relational facts.}
\begin{tabular*}{0.95\linewidth}{@{}cccccc@{}}
\bottomrule
\multicolumn{1}{c}{\textbf{Datasets}} &\multicolumn{1}{c}{\textbf{Train}} &\multicolumn{1}{c}{\textbf{Valid}} &\multicolumn{1}{c}{\textbf{Test}} &\multicolumn{1}{c}{\textbf{Entity}} &\multicolumn{1}{c}{\textbf{Relation}}\\
\hline
WD50K  &166435  &23913  &46159  &47155  &531 \\
WD50K (33)  &73406  &10568  &18133  &38123  &474 \\
WD50K (66)  &35968  &5154  &8045  &27346  &403 \\
WD50K (100) &22738  &3279  &5297  &18791  &278 \\
\hline
WikiPeople  &294439  &37715  &37712  &34825  &178 \\
WikiPeople (33)  &28280  &3550  &3542  &20921  &145 \\
WikiPeople (66)  &14130  &1782  &1774  &13651  &133 \\
WikiPeople (100)  &9319  &1181  &1173  &8068  &105 \\
\hline
JF17K   &76379  &-  &24568  &28645  &501 \\
JF17K (33)  &56959  &8122  &9112  &24081  &490 \\
JF17K (66)  &27280  &4413  &5403  &19288  &469 \\
JF17K (100)  &17190  &3152  &4142  &12656  &307 \\
\bottomrule
\end{tabular*}
\label{table_statistics_datasets_scenarios_1}
\end{table}

\vic{In this section, we present the results of the conducted experiments. We first describe the datasets, \zhiweihu{evaluation protocol and implementation details} (\S~\ref{section_datasets_and_baselines}), and describe the baseline models (\S~\ref{section_baselines}). We then discuss the \zhiwei{main} experimental results (\S.~\ref{section_main_results}). Finally, we present results of our ablation \zhiwei{studies} (\S~\ref{section_ablation_study}).}

\subsection{\zhiweihu{Experiment Setup}}
\label{section_datasets_and_baselines}
\subsubsection{\zhiweihu{Datasets.}} 
\vic{We evaluate HyperFormer on three \zhiweihu{well-known} datasets: WD50K~\citep{Mikhail_2020}, WikiPeople~\citep{Saiping_2019}, and JF17K~\citep{Jianfeng_2016}. \zhiweihu{WD50K and WikiPeople are derived from Wikidata, and JF17K is collected from Freebase.} These datasets have the following two characteristics: i) only certain percentage of main triples contain qualifiers, 13.6\% in WD50K, 2.6\% in WikiPeople and 45.9\% in JF17K; ii) each triple contains a different number of qualifiers, 0\textasciitilde7 for WikiPeople, and 0\textasciitilde4 for JF17K, where the qualifier  number  means that the main triple does not contain hyper-relational knowledge. We refine these datasets from two perspectives, based on the percentage of triples containing hyper-relational knowledge and on the  number of qualifiers associated to triples. So, we construct three datasets with different conditions: \emph{Mixed-percentage Mixed-qualifier, Fixed-percentage Mixed-qualifier, Fixed-percentage Fixed-qualifier}, where Mixed-percentage and Mixed-qualifier respectively  indicate that the number of triples with qualifiers is arbitrary (not fixed) and that the number of qualifiers per triple is not fixed. The Fixed condition is defined as expected, and 
clearly, there is no Mixed-percentange in the Fixed qualifier scenario.
In addition, we also construct the datasets in which all entities have  low degree. The four scenarios are specifically described as follows:}
\begin{enumerate}[itemsep=0.5ex, leftmargin=5mm]
\item 
\textbf{Mixed-percentage Mixed-qualifier.} These datasets  are directly taken from~\citep{Mikhail_2020}. They aim  at  verifying the generalization performance in the scenario where
the percentage of triples with qualfiers and the number of qualifiers associated with each triple is arbitrary.
\item 
\vic{\textbf{Fixed-percentage Mixed-qualifier.} We construct subsets of existing datasets in which the  percentage of triples with qualifiers is fixed.  For example, for WD50K we construct: WD50K (33), WD50K (66) and WD50K (100), with the number in parentheses representing the percentage of triples with qualifiers. We construct similar subsets for the WikiPeople and JF17K datasets. \zhiweihu{The corresponding datasets statistics are presented in Table~\ref{table_statistics_datasets_scenarios_1}.}}
\item \zhiweihu{\textbf{Fixed-percentage Fixed-qualifier.} Due to the sparsity of higher qualifier facts in WikiPeople and JF17K datasets, we follow GETD~\citep{Yu_2020} to filter out the triples with 3 and 4 associated qualifiers, obtaining  WikiPeople-3, WikiPeople-4, JF17K-3, and JF17K-4, respectively. The corresponding datasets statistics are presented in Table~\ref{table_statistics_datasets_scenarios_2}. 
}
\item \vic{\zhiweihu{\textbf{Entities with Low Degree.}} To evaluate the performance of the tested models depending on the  node degrees (number of neighbors), we construct  subsets of existing datasets in which all nodes have the low degree. In this case, we select as the basic datasets data in which \emph{all} triples have qualifiers.  For example, for  WD50K (100), we construct four subsets with node degrees from one to four, denoted as WD50K (100) \#1, WD50K (100) \#2, WD50K (100) \#3, and WD50K (100) \#4. \zhiweihu{The corresponding datasets statistics are presented in Table~\ref{table_statistics_datasets_scenarios_3}.}}

\end{enumerate}

\begin{table}[!htp]
\setlength{\abovecaptionskip}{0.03cm}
\renewcommand\arraystretch{1.2}
\setlength{\tabcolsep}{0.58em}
\centering
\small
\caption{\zhiweihu{Statistics of datasets under fixed-percentage fixed-qualifier scenarios}.}
\begin{tabular*}{0.85\linewidth}{@{}cccccc@{}}
\bottomrule
\multicolumn{1}{c}{\textbf{Datasets}} &\multicolumn{1}{c}{\textbf{Train}} &\multicolumn{1}{c}{\textbf{Valid}} &\multicolumn{1}{c}{\textbf{Test}} &\multicolumn{1}{c}{\textbf{Entity}} &\multicolumn{1}{c}{\textbf{Relation}}\\
\hline
WikiPeople-3  &20656  &2582  &2582  &12270  &66 \\
WikiPeople-4  &12150  &1519  &1519  &9528  &50 \\
JF17K-3  &27635  &3454  &3455  &11541  &104 \\
JF17K-4  &7607  &951  &951  &6536  &23 \\
\bottomrule
\end{tabular*}
\label{table_statistics_datasets_scenarios_2}
\end{table}

\begin{table}[!htp]
\setlength{\abovecaptionskip}{0.03cm}
\renewcommand\arraystretch{1.2}
\setlength{\tabcolsep}{0.58em}
\centering
\small
\caption{Statistics of datasets \zhiweihu{with different number of node degrees}. The value behind \# indicates that the entity in the training set only contains the number of neighbors with the corresponding value. }
\begin{tabular*}{0.95\linewidth}{@{}cccccc@{}}
\bottomrule
\multicolumn{1}{c}{\textbf{Datasets}} &\multicolumn{1}{c}{\textbf{Train}} &\multicolumn{1}{c}{\textbf{Valid}} &\multicolumn{1}{c}{\textbf{Test}} &\multicolumn{1}{c}{\textbf{Entity}} &\multicolumn{1}{c}{\textbf{Relation}}\\
\hline
WD50K (100) \#1  &2191  &3279  &5297  &10375  &189 \\
WD50K (100) \#2  &4382  &3279  &5297  &11241  &200 \\
WD50K (100) \#3  &6547  &3279  &5297  &11985  &207 \\
WD50K (100) \#4  &8506  &3279  &5297  &12649  &210 \\
\hline
WikiPeople (100) \#1  &1253  &1181  &1173  &4212  &83 \\
WikiPeople (100) \#2  &2498  &1181  &1173  &4711  &85 \\
WikiPeople (100) \#3  &3647  &1181  &1173  &5040  &87 \\
WikiPeople (100) \#4  &4515  &1181  &1173  &5338  &89 \\
\hline
JF17K (100) \#1  &2492  &3152  &4142  &7320  &253 \\
JF17K (100) \#2  &4984  &3152  &4142  &7930  &255 \\
JF17K (100) \#3  &7294  &3152  &4142  &8367  &257 \\
JF17K (100) \#4  &9219  &3152  &4142  &8688  &259 \\
\bottomrule
\end{tabular*}
\label{table_statistics_datasets_scenarios_3}
\end{table}

\subsubsection{\zhiweihu{Baselines}}
\label{section_baselines}
We compare HyperFormer with various state-of-the-art methods for hyper-relational knowledge graph completion: m-TransH~\citep{Jianfeng_2016}, RAE~\citep{Richong_2018}, NaLP-Fix~\citep{Paolo_2020}, HINGE~\citep{Paolo_2020}, StarE~\citep{Mikhail_2020}, Hy-Transformer~\citep{Donghan_2021}, GRAN~\citep{Quan_2021}, and QUAD~\citep{Harry_2022}. Note that GRAN contains three variants, i.e., GRAN{\small{-hete}}, GRAN{\small{-homo}} and GRAN{\small{-complete}}. If there is no special suffix,   GRAN denotes GRAN{\small{-hete}}. There are two variants of QUAD: QUAD and QUAD (Parallel). If there is no special suffix, QUAD denotes QUAD (Parallel). 

\subsubsection{\zhiweihu{Evaluation Protocol.}}
\zhiweihu{We evaluate the model performance using two common metrics: MRR and Hits@\emph{N} (abbreviated as H@\emph{N}). MRR is the average of reciprocal ranking, and Hits@\emph{N} is the proportion of top \emph{N} (we use \emph{N}=\{1,3,10\}). For both metrics, the larger the value, the better the performance of a model.}

\subsubsection{\zhiweihu{Implementation Details.}}
\zhiweihu{All experiments are conducted on six 32G Tesla V100 GPUs. Our method is implemented with PyTorch. We employ AdamW~\citep{adamw_2019} as the optimizer and a cosine decay scheduler with linear warm-up is used for optimization. We determine the hyperparameter values by using a grid search based on the MRR performance on the validation dataset. We select the neighbors of an entity in \{2, \textbf{3}, 4\}, the qualifier pairs of a relation in \{5, \textbf{6}, 7\}, the learning rate in \{3e-4, 4e-4, 5e-4, \textbf{6e-4}, 7e-4\}, the label smoothing factor in \{0.3, 0.5, 0.7, \textbf{0.9}\}, the  number of layer in a Transformer in \{2, 4, \textbf{8}, 16\}, the head number in \{1, \textbf{2}, 4, 8\}, the Transformer input dropout rate in \{0.6, \textbf{0.7}, 0.8\}, the Transformer hidden dropout rate in \{\textbf{0.1}, 0.2, 0.3\}, the dimensions of the embedding size in \{80, 200, 320, \textbf{400}\}, the number of convolution channels in \{64, \textbf{96}, 128\}, the convolutional kernel size is \textbf{9}, the convolutional input dropout rate in \{0.1, \textbf{0.2}, 0.3\}, the convolutional hidden dropout rate in \{0.4, \textbf{0.5}, 0.6\}, the number of experts in the MoE module in \{8, 16, 32, \textbf{64}\}, the number of top experts in the MoE module in \{\textbf{2}, 4, 6, 8\}.}

\begin{table*}[!htp]
\setlength{\abovecaptionskip}{0.05cm}
\renewcommand\arraystretch{1.3}
\setlength{\tabcolsep}{0.36em}
\centering
\small
\caption{Evaluation of different models \zhiweihu{with mixed-percentage mixed-qualifier} on the WD50K, WikiPeople and JF17K datasets. All baseline results are collected from the original literature. Best scores are highlighted in \textbf{bold}, the second best scores are \uline{underlined}, and ’--’ indicates the results are not reported in previous work.}
\begin{tabular*}{0.61\linewidth}{@{}cccccccccc@{}}
\bottomrule
\multicolumn{1}{c}{\multirow{2}{*}{\textbf{Methods}}} & \multicolumn{3}{c}{\textbf{WD50K (13.6)}} & \multicolumn{3}{c}{\textbf{WikiPeople (2.6)}} & \multicolumn{3}{c}{\textbf{JF17K (45.9)}}\\
\cline{2-4}\cline{5-7}\cline{8-10}
& \textbf{MRR} & \textbf{H@1} & \textbf{H@10} & \textbf{MRR} & \textbf{H@1} & \textbf{H@10} & \textbf{MRR}  & \textbf{H@1} & \textbf{H@10} \\
\hline
m-TransH~\citep{Jianfeng_2016}   &--  &--  &--  &0.063  &0.063  &0.300  &0.206  &0.206    &0.463 \\
RAE~\citep{Richong_2018}   &--  &--  &--  &0.059  &0.059  &0.306  &0.215  &0.215    &0.469 \\
NaLP-Fix~\citep{Paolo_2020}   &0.177  &0.131  &0.264  &0.420  &0.343  &0.556  &0.245  &0.185    &0.358 \\
HINGE~\citep{Paolo_2020}    &0.243  &0.176  &0.377  &0.476  &0.415  &0.585  &0.449  &0.361    &0.624 \\
StarE~\citep{Mikhail_2020}    &0.349  &0.271  &0.496  &0.491  &0.398  &\textbf{0.648}  &0.574  &0.496    &0.725 \\
Hy-Transformer~\citep{Donghan_2021}  &\uline{0.356}  &\uline{0.281}  &\uline{0.498}  &\uline{0.501}  &0.426  &0.634  &0.582  &0.501    &0.742 \\
GRAN{\small{-homo}}~\citep{Quan_2021}   &--  &--  &--  &0.487  &0.410  &0.618  &0.611  &0.533    &0.767 \\
GRAN{\small{-complete}}~\citep{Quan_2021}   &--  &--  &--  &0.489  &0.413  &0.617  &0.591  &0.510    &0.753 \\
GRAN{\small{-hete}}~\citep{Quan_2021}   &--  &--  &--  &\textbf{0.503}  &\textbf{0.438}  &0.620  &\uline{0.617}  &\uline{0.539}    &\uline{0.770} \\
QUAD~\citep{Harry_2022}   &0.348  &0.270  &0.497  &0.466  &0.365  &0.624  &0.582  &0.502    &0.740 \\
QUAD (Parallel)~\citep{Harry_2022}    &0.349  &0.275  &0.489  &0.497  &\uline{0.431}  &0.617  &0.596  &0.519    &0.751 \\
HyperFormer    &\textbf{0.366}  &\textbf{0.288}  &\textbf{0.514}  &0.473  &0.361  &\uline{0.646}  &\textbf{0.664}  &\textbf{0.601}    &\textbf{0.787} \\
\bottomrule
\end{tabular*}
\label{table_main_result}
\end{table*}

\begin{table*}[!htp]
\setlength{\abovecaptionskip}{0.05cm}
\renewcommand\arraystretch{1.3}
\setlength{\tabcolsep}{0.30em}
\centering
\small
\caption{Evaluation of different models \zhiweihu{with fixed-percentage mixed-qualifier} on the WD50K, WikiPeople and JF17K datasets. Best scores are highlighted in \textbf{bold}.}
\begin{tabular*}{0.99\linewidth}{@{}ccccccccccccccccccc@{}}
\bottomrule
\multicolumn{1}{c}{\multirow{3}{*}{\textbf{Methods}}} & \multicolumn{6}{c}{\textbf{WD50K}} & \multicolumn{6}{c}{\textbf{WikiPeople}} & \multicolumn{6}{c}{\textbf{JF17K}}\\
\cline{2-7}\cline{8-13}\cline{14-19}
& \multicolumn{2}{c}{\textbf{33\%}} & \multicolumn{2}{c}{\textbf{66\%}} & \multicolumn{2}{c}{\textbf{100\%}} & \multicolumn{2}{c}{\textbf{33\%}} & \multicolumn{2}{c}{\textbf{66\%}} & \multicolumn{2}{c}{\textbf{100\%}} & \multicolumn{2}{c}{\textbf{33\%}}  & \multicolumn{2}{c}{\textbf{66\%}} & \multicolumn{2}{c}{\textbf{100\%}} \\
\cline{2-3}\cline{4-5}\cline{6-7}\cline{8-9}\cline{10-11}\cline{12-13}\cline{14-15}\cline{16-17}\cline{18-19}

& \textbf{MRR} & \textbf{H@1} & \textbf{MRR} & \textbf{H@1} & \textbf{MRR}  & \textbf{H@1} & \textbf{MRR} & \textbf{H@1} & \textbf{MRR} & \textbf{H@1} & \textbf{MRR}  & \textbf{H@1} & \textbf{MRR} & \textbf{H@1} & \textbf{MRR} & \textbf{H@1} & \textbf{MRR}  & \textbf{H@1} \\

\hline
StarE~\citep{Mikhail_2020}   &0.308  &0.247  &0.449  &0.388  &0.610  &0.543    &0.192  &0.143 &0.259 &0.205 &0.343 &0.279   &0.290 &0.197   &0.302 &0.214   &0.321 &0.223\\
Hy-Transformer~\citep{Donghan_2021}   &0.313  &0.255  &0.458  &0.397  &0.621  &0.557    &0.192  &0.140 &0.268 &0.215 &0.372 &0.316   &0.298 &0.204   &0.325 &0.234   &0.361 &0.266\\
GRAN~\citep{Quan_2021}   &0.322  &0.269  &0.472  &0.419  &0.647  &0.593    &0.201  &0.156 &0.287 &0.244 &0.403 &0.349   &0.307 &0.212   &0.326 &0.237   &0.382 &0.290\\
QUAD~\citep{Harry_2022}   &0.329  &0.266  &0.479  &0.416  &0.646  &0.572    &0.204  &0.155 &0.282 &0.228 &0.385 &0.318   &0.307 &0.210   &0.334 &0.241   &0.379 &0.277\\
HyperFormer   &\textbf{0.338}  &\textbf{0.280}  &\textbf{0.492}  &\textbf{0.434}  &\textbf{0.666}  &\textbf{0.611}    &\textbf{0.213}  &\textbf{0.161} &\textbf{0.298} &\textbf{0.255} &\textbf{0.426} &\textbf{0.373}   &\textbf{0.352} &\textbf{0.254}   &\textbf{0.411} &\textbf{0.325}   &\textbf{0.478} &\textbf{0.396} \\

\bottomrule
\end{tabular*}
\label{table_mrr_different_hyper_relation_ratio}
\end{table*}

\subsection{\zhiwei{Main} Results}
\label{section_main_results}
\zhiweihu{\noindent\textbf{Mixed-percentage Mixed-qualifier.}}
\vic{Table~\ref{table_main_result} reports the results 
on the Hyper-relational KGC task \zhiweihu{with mixed-percentage mixed-qualifier} on the WD50K, WikiPeople, and JF17K datasets. We can observe that  HyperFormer significantly outperforms all baselines on the WD50K and JF17K datasets. Specifically, HyperFormer respectively achieves performance improvements of 1.0\% / 0.7\% / 1.6\% in MRR / Hits@1 / Hits@10 on WD50K, compared to the best performing baseline, \zhiweihu{Hy-Transformer}. It gets analogous improvements of 4.7\% / 6.2\% / 1.7\% on JF17K. On WikiPeople its performance is slightly below the SoTA. These results can be explained by the fact that both WD50K and JF17K contain a relatively high percentage of triples with qualifier pairs: 13.6\% and 45.9\%, respectively. However,  WikiPeople has a much lower percentage of triples with qualifiers, 2.6\%, so the triple-only facts dominate the overall score. 
 Hyperformer successfully exploits the interaction between entities, relations and \zhiweihu{qualifiers}  to improve the performance on the HKGC task, especially on datasets with a rich amount of hyper-relational knowledge.} \\
\zhiweihu{\noindent\textbf{Fixed-percentage Mixed-qualifier.}}
\vic{We also investigate the effectiveness of HyperFormer and the baselines under different ratios of relational facts with qualifiers. 
For each of the used datasets, we obtained three subsets (as described in Point~2 in Section~\ref{section_datasets_and_baselines}) containing approximately 33\%, 66\%, and 100\%  of  facts with qualifiers.} \vic{Table~\ref{table_mrr_different_hyper_relation_ratio} presents an overview of the obtained results. We observe that HyperFormer gets larger improvements over the baselines when the percentage of available facts with qualifiers is higher. Specifically, on WikiPeople,  HyperFormer  respectively achieves improvements over QUAD of 0.9\% / 1.6\% / 4.1\% in  the 33\% / 66\% / 100\% variants.} \vic{This shows that an important reason of why Hyperformer could not surpass GRAN-hete in the mixed-percentage mixed-qualifier HKGC task on the WikiPeople dataset (cf. Table~\ref{table_main_result}) is that it only contains a very small amount of triples with  hyper-relational knowledge.} Indeed, the main strength of Hyperformer is in the integration of hyper-relational knowledge by capturing the interaction of entities, relations and qualifiers.\\
\zhiweihu{\noindent\textbf{Fixed-percentage Fixed-qualifier.}}
\zhiweihu{We   investigate the performance on hyper-relational data with fixed number of qualifiers in Table \ref{table_different_n_ary}. We   observe that 
HyperFormer consistently achieves state-of-the-art performance on 
all datasets in Table \ref{table_different_n_ary}. 
At the same time, we find out that the performance of all models is significantly lower in the mixed-percentage mixed-qualifier datasets than in scenarios with a fixed number of hyper-relational knowledge, which is consistent for both WikiPeople and JF17K. This might be explained by the fact that uneven distributions among different quantities of hyper-relational facts may affect the stability of model training.
} 
\begin{table*}[!htp]
\setlength{\abovecaptionskip}{0.05cm}
\renewcommand\arraystretch{1.3}
\setlength{\tabcolsep}{0.3em}
\centering
\small
\caption{\zhiweihu{Evaluation of different models with fixed-percentage fixed-qualifier on WikiPeople and JF17K datasets. Best scores are highlighted in \textbf{bold}. }}
\begin{tabular*}{0.93\linewidth}{@{}ccccccccccccccccc@{}}
\bottomrule
\multicolumn{1}{c}{\multirow{2}{*}{\textbf{Methods}}}   & \multicolumn{4}{c}{\textbf{WikiPeople-3}} & \multicolumn{4}{c}{\textbf{WikiPeople-4}} & \multicolumn{4}{c}{\textbf{JF17K-3}} & \multicolumn{4}{c}{\textbf{JF17K-4}}\\
\cline{2-5}\cline{6-9}\cline{10-13}\cline{14-17}

& \textbf{MRR} & \textbf{H@1}   & \textbf{H@3} & \textbf{H@10} & \textbf{MRR} & \textbf{H@1} & \textbf{H@3} & \textbf{H@10} & \textbf{MRR}  & \textbf{H@1} & \textbf{H@3}   & \textbf{H@10} & \textbf{MRR} & \textbf{H@1} & \textbf{H@3}    & \textbf{H@10}\\
\hline

StarE~\citep{Mikhail_2020}    &0.401  &0.310   &0.434  &0.592  &0.243  &0.156  &0.269  &0.430  &0.707  &0.635  &0.744    &0.847  &0.723  &0.669    &0.753  &0.839 \\

Hy-Transformer~\citep{Donghan_2021}    &0.403  &0.323   &0.436  &0.569  &0.248  &0.165  &0.275  &0.422  &0.690  &0.617  &0.725    &0.837  &0.773  &0.717    &0.806  &0.875 \\

GRAN~\citep{Quan_2021}    &0.397  &0.328   &0.429  &0.533  &0.239  &0.178  &0.261  &0.364  &0.779  &0.724  &0.811    &0.893  &0.798  &0.744    &0.830  &0.904 \\

QUAD~\citep{Harry_2022}    &0.403  &0.321   &0.438  &0.563  &0.251  &0.167  &0.280  &0.425  &0.730  &0.660  &0.767    &0.870  &0.787  &0.730    &0.823  &0.895 \\

HyperFormer   &\textbf{0.573}  &\textbf{0.511}  &\textbf{0.603}  &\textbf{0.693}  &\textbf{0.393} &\textbf{0.336}  &\textbf{0.415} &\textbf{0.496}  &\textbf{0.832}  &\textbf{0.790}   &\textbf{0.855}    &\textbf{0.914}  &\textbf{0.857}  &\textbf{0.811}  &\textbf{0.884}   &\textbf{0.937} \\

\bottomrule
\end{tabular*}
\label{table_different_n_ary}
\end{table*}

\begin{table*}[!htp]
\setlength{\abovecaptionskip}{0.05cm}
\renewcommand\arraystretch{1.2}
\setlength{\tabcolsep}{0.32em}
\centering
\small
\caption{MRR results of different node degrees on the WD50K(100), WikiPeople(100) and JF17K(100) datasets. The last line shows the difference between \textbf{best} scores and the \uline{second best} scores.}
\begin{tabular*}{0.75\linewidth}{@{}ccccccccccccc@{}}
\bottomrule
\multicolumn{1}{c}{\multirow{2}{*}{\textbf{Methods}}} & \multicolumn{4}{c}{\textbf{WD50K (100)}} & \multicolumn{4}{c}{\textbf{WikiPeople (100)}} & \multicolumn{4}{c}{\textbf{JF17K (100)}}\\
\cline{2-5}\cline{6-9}\cline{10-13}
& \textbf{\#1} & \textbf{\#2} & \textbf{\#3}   & \textbf{\#4} & \textbf{\#1} & \textbf{\#2} & \textbf{\#3} & \textbf{\#4} & \textbf{\#1}  & \textbf{\#2} & \textbf{\#3} & \textbf{\#4} \\
\bottomrule
StarE~\citep{Mikhail_2020}   &0.104  &0.208  &0.313  &0.369  &\uline{0.121}  &0.112  &0.193  &0.255    &0.169  &0.249  &0.275  &0.286 \\
Hy-Transformer~\citep{Donghan_2021}   &0.071  &0.167  &0.315  &\uline{0.374}  &0.091  &0.148  &0.186  &0.233    &0.137  &0.241  &\uline{0.299}  &\uline{0.318} \\
GRAN~\citep{Quan_2021}   &\uline{0.125}  &\uline{0.235}  &\uline{0.327}  &\uline{0.374}  &0.119  &\uline{0.186}  &\uline{0.242}  &\uline{0.273}    &0.203  &\uline{0.267}  &0.284  &0.301 \\
QUAD~\citep{Harry_2022}   &0.065  &0.134  &0.284  &0.371  &0.075  &0.140  &0.186  &0.255    &\uline{0.228}  &0.241  &0.280  &0.306 \\
HyperFormer   &\textbf{0.193}  &\textbf{0.303}  &\textbf{0.374}  &\textbf{0.410}  &\textbf{0.194}  &\textbf{0.252}  &\textbf{0.303}  &\textbf{0.328}    &\textbf{0.305}  &\textbf{0.338}  &\textbf{0.350}  &\textbf{0.374} \\
\bottomrule
Absolute improvement (\%)   &6.8\%  &6.8\%  &4.7\%  &3.6\%  &7.3\%  &6.6\%  &6.1\%  &5.5\%    &7.7\%  &7.1\%  &5.1\%  &5.6\% \\
\bottomrule
\end{tabular*}
\label{table_mrr_different_degree}
\end{table*}
\noindent
\zhiweihu{\textbf{Different Numbers of Neighbors.}}
\vic{We also investigate the performance of the models on entities \vi{with few neighbors} in Table \ref{table_mrr_different_degree}. In this case we look at training datasets in which all entities have one, two, three or four neighbors  (see Point~4 in Section~\ref{section_datasets_and_baselines}), while the validation and test sets remain unchanged. We found that the baseline models perform very poorly on these subsets. For example, in WD50K\_100 (\#1), StarE, and QUAD can respectively obtain an MRR metric of 10.4\% and 6.5\%, while HyperFormer can achieve 19.3\%. \zhiweihu{This is explained by the way that these two models use the global-level structure to encode qualifier knowledge into the relation representation, which is suitable for scenarios in which all nodes have several neighbors. GRAN achieves 12.5\%, since it does encode qualifiers into the main triple in a local-level fashion like HyperFormer, but it  ignores the structural content of qualifier pairs.} Differently, HyperFormer proposes a new  integration method that realizes the interaction between entities, relations and qualifiers. } 

\subsection{Ablation Studies}
\begin{figure*}[htbp]
        \centering
        \subcaptionbox{Different variants of model}{
        \includegraphics[width = .24\linewidth]{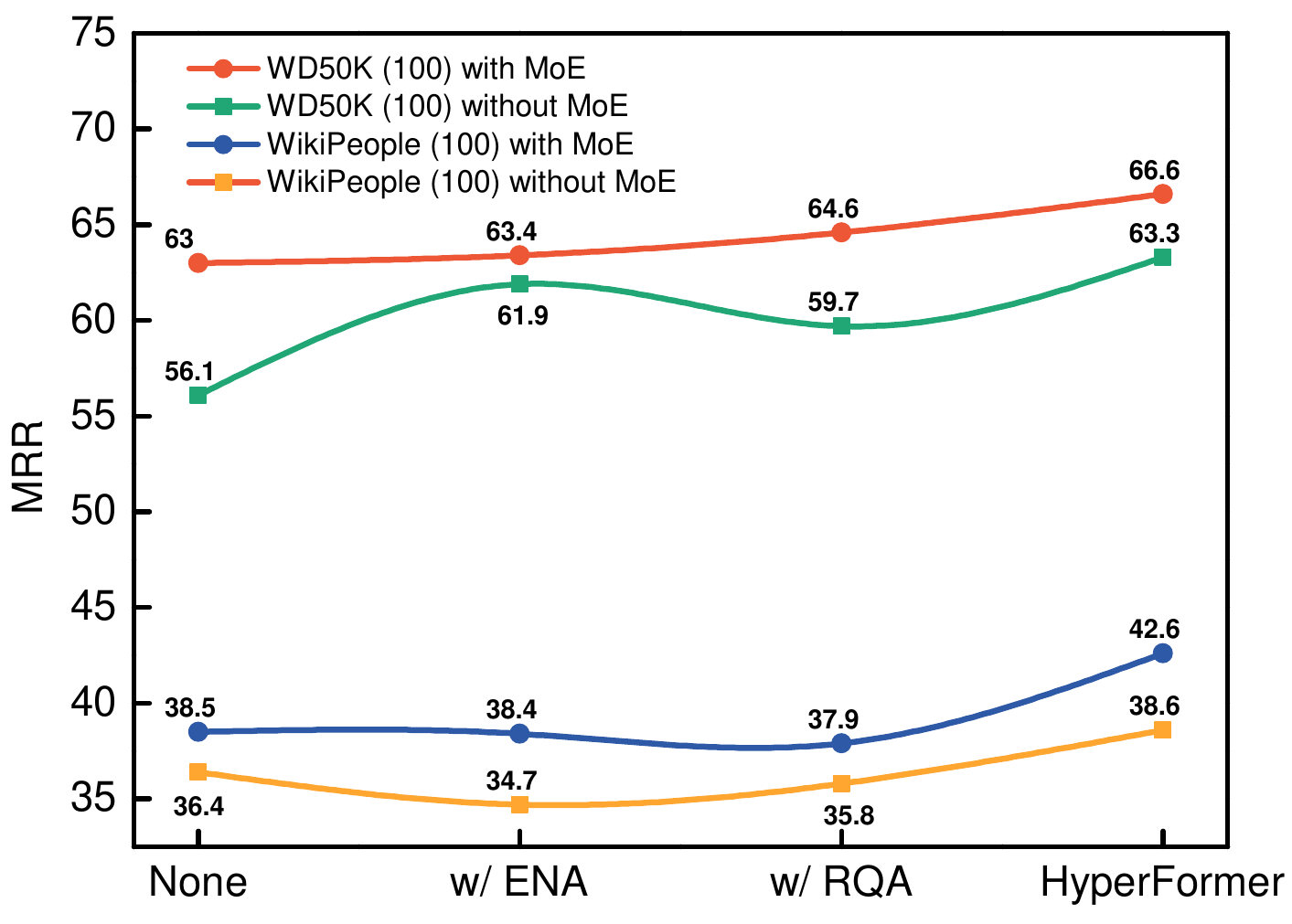}
        }
        \subcaptionbox{Different hidden sizes}{
        \centering
        \includegraphics[width = .23\linewidth]{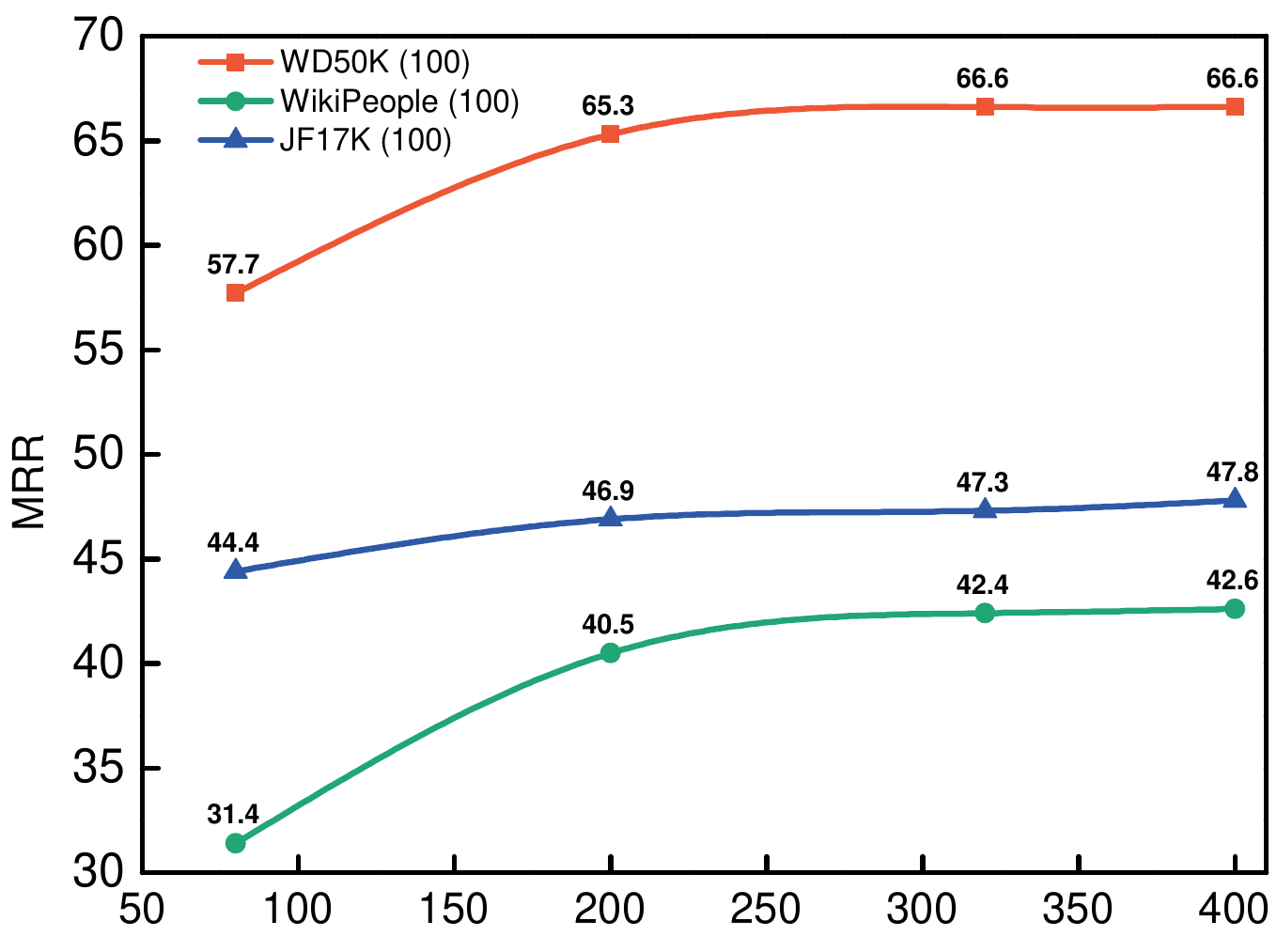}
        }
        \subcaptionbox{Number of experts}{
        \centering
        \includegraphics[width = .23\linewidth]{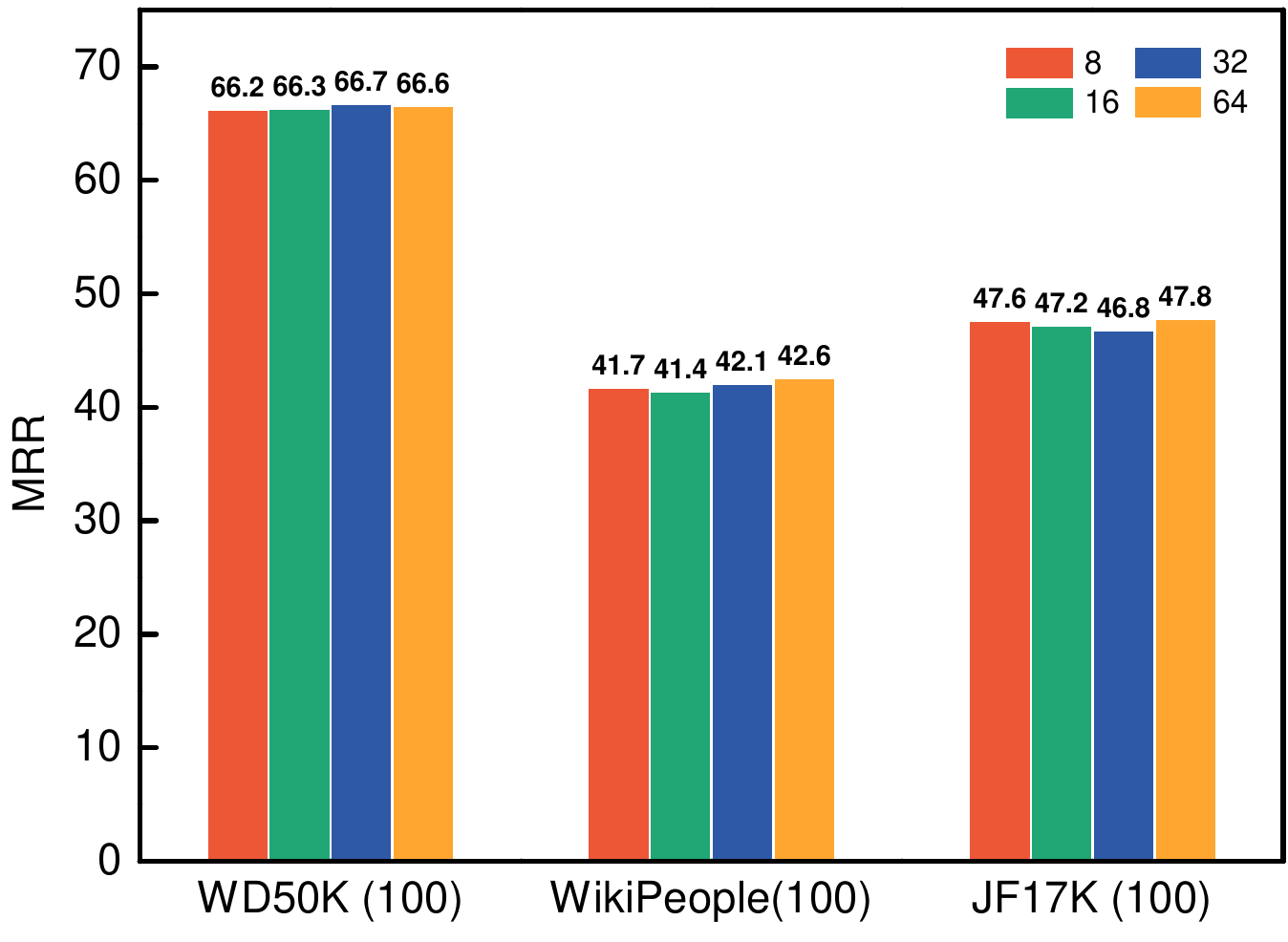}
        }
        \subcaptionbox{Label smoothing}{
        \centering
        \includegraphics[width = .23\linewidth]{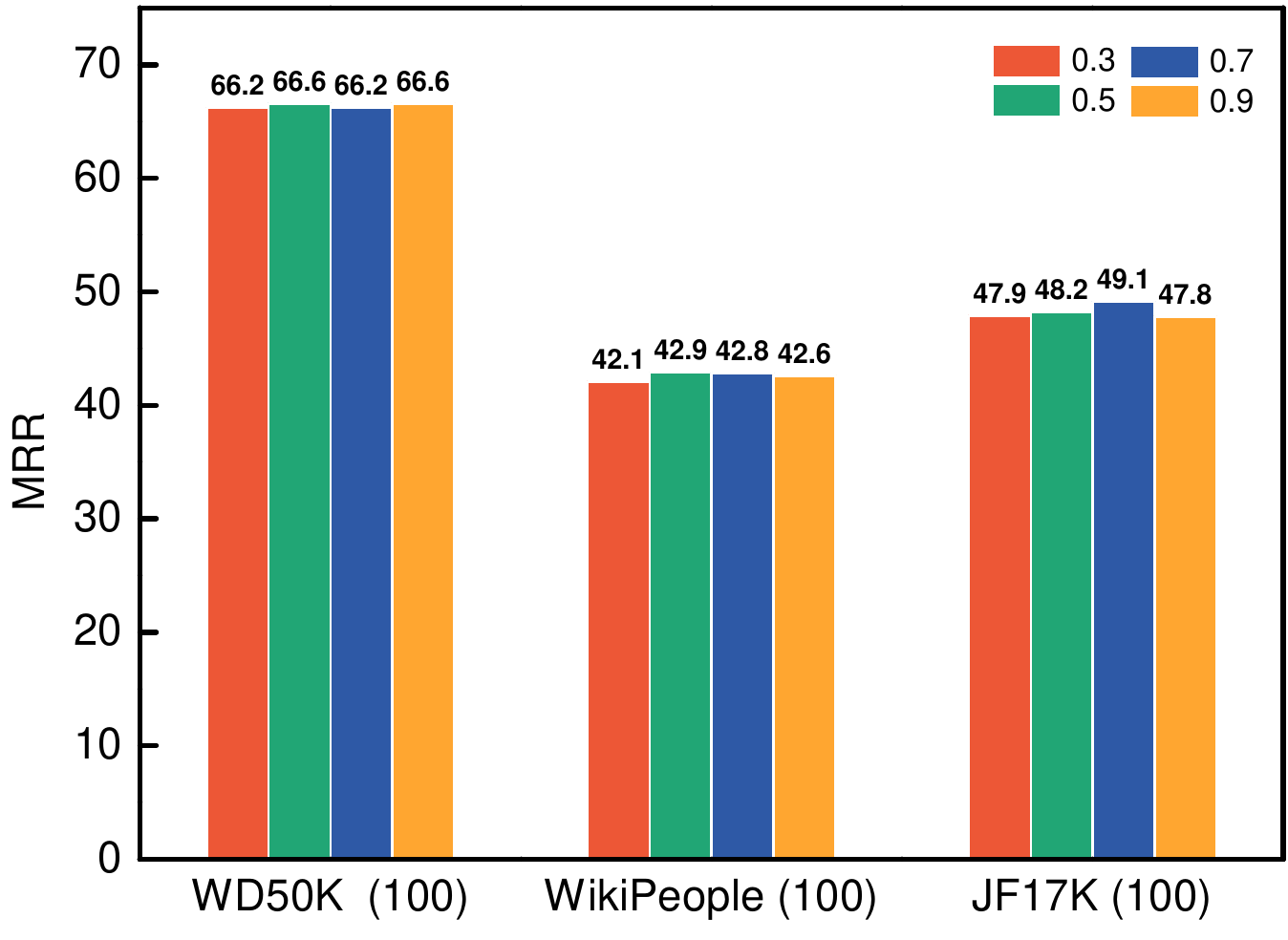}
        }
        \caption{The ablation studies results under different experimental conditions.}
        \label{figure_ablation_studies}
\end{figure*}

\vic{We verify the contribution of each component of HyperFormer and the effect of different hyperparameters on the performance. \zhiweihu{First, we explore the impact of different translation operations on the performance, cf. Table~\ref{table_different_translation_methods}.} \zhiweihu{Then,} we look at different variants of the model, and different hidden sizes, number of experts, and  values of label smoothing, \zhiweihu{cf. Figure~\ref{figure_ablation_studies}}. Finally, we show in Table~\ref{table_flops_and_params} the amount of parameters and calculations with and without MoE.} \\
\zhiweihu{\noindent\textbf{Different Translation Methods.}}
\zhiweihu{
Table~\ref{table_different_translation_methods} shows  detailed results of selecting different translation methods to compose the qualifier entity and qualifier relation. Specifically, we adopt four translation methods, i.e., TransE~\citep{Antoine_2013}, DistMult~\citep{Bishan_2015}, ComplEx~\citep{TrouillonWRGB_2016}, and RotatE~\citep{Zhiqing_2019}. We find that the selection of translation methods has little impact on the performance. This may be because different translation methods are only used to convert entities and relations in the qualifier pairs into single vector, while the subsequent CBI module is used to determine the combined representation of qualifier pairs and its impact on entities and relations.
} \\
\zhiweihu{\noindent\textbf{Different Variants of HyperFormer.}}
\vic{Figure~\ref{figure_ablation_studies}(a) presents the results on the impact of each component of HyperFormer. We consider four variants with/without MoE in  the transformer part: (i)  without any modules, denoted  \emph{None}; (ii)   with the ENA module as the only component, denoted \emph{w/ ENA}; (iii) with the RQA module as the only component, denoted  \emph{w/ RQA}; (iv)  with both ENA, RQA, and CBI, denoted as HyperFormer. We observe that the introduction of the ENA or RQA module in some cases can bring an improvement in the performance compared to \emph{None}. The addition of the CBI module brings a consistent improvement, because CBI  realizes the bidirectional interaction between entities, relations, and hyper-relational knowledge. In addition, after adding the MoE mechanism to the transformer, a further improvement is obtained.} \\
\zhiweihu{\noindent\textbf{Different Hidden Sizes.}}
\vic{Figure~\ref{figure_ablation_studies}(b) presents results showing the influence of the hidden sizes. We observe that the increase of the hidden size helps to capture a large amount of messages, which can improve the model performance. However, after the hidden size is set to 320, the results show a stable trend. This indicates that it is not the case that the higher the embedding dimension, the better the performance of the model, which is consistent with the finding by~\citep{Xiao_2022}. Note that setting a larger hidden size requires more video memory, so in practice,  after the performance is stable the minimum value is set as the final one.}
\begin{table*}[!htp]
\setlength{\abovecaptionskip}{0.03cm}
\renewcommand\arraystretch{1.35}
\setlength{\tabcolsep}{0.3em}
\centering
\small
\caption{Evaluation of different transaction methods on WD50K(100), WikiPeople(100) and JF17K(100) datasets. Best scores are highlighted in \textbf{bold}.}
\begin{tabular*}{0.75\linewidth}{@{}ccccccccccccc@{}}
\bottomrule
\multicolumn{1}{c}{\multirow{2}{*}{\textbf{Methods}}} & \multicolumn{4}{c}{\textbf{WD50K (100)}} & \multicolumn{4}{c}{\textbf{WikiPeople (100)}} & \multicolumn{4}{c}{\textbf{JF17K (100)}}\\
\cline{2-5}\cline{6-9}\cline{10-13}
& \textbf{MRR} & \textbf{H@1} & \textbf{H@3}   & \textbf{H@10} & \textbf{MRR} & \textbf{H@1} & \textbf{H@3} & \textbf{H@10} & \textbf{MRR}  & \textbf{H@1} & \textbf{H@3} & \textbf{H@10} \\
\hline
HyperFormer-TransE   &0.666  &\textbf{0.611}  &0.697  &0.768  &0.424  &\textbf{0.374}  &0.452  &0.524    &\textbf{0.487}  &\textbf{0.401}  &\textbf{0.526}  &\textbf{0.662} \\
HyperFormer-DistMult   &0.666  &\textbf{0.611}  &\textbf{0.698}  &0.770  &\textbf{0.426}  &0.373  &\textbf{0.454}  &\textbf{0.527}    &0.478  &0.396  &0.515  &0.645 \\
HyperFormer-ComplEx  &\textbf{0.667}  &\textbf{0.611}  &\textbf{0.698}  &0.769  &0.422  &0.370  &0.445  &0.518    &0.479  &0.399  &0.519  &0.642 \\
HyperFormer-RotatE  &0.655  &0.592  &0.690  &\textbf{0.772}  &0.415  &0.371  &0.434  &0.496    &0.479  &\textbf{0.401}  &0.515  &0.644 \\
\bottomrule
\end{tabular*}
\label{table_different_translation_methods}
\end{table*}
\\
\noindent
\zhiweihu{\textbf{Number of Experts.}}
\vic{We also investigate the impact of the number of experts on the  performance of HyperFormer, cf. Figure~\ref{figure_ablation_studies}(c). The  selection of  the number of experts is an important factor  as it directly affects  the size of the model. We observe that by selecting a larger number of experts a slight improvement is obtained. In practice, it is necessary to balance the number of video memory and experts, while ensuring that the number of experts remains as small as possible without  affecting the performance.}
\begin{table}[!htp]
\setlength{\abovecaptionskip}{0.03cm}
\renewcommand\arraystretch{1.2}
\setlength{\tabcolsep}{0.3em}
\centering
\small
\caption{\zwh{The amount of parameters and calculation with or without the introduction of MoE. w/ means with, w/o means without.}}
\begin{tabular*}{0.8\linewidth}{@{}ccccc@{}}
\bottomrule
\multicolumn{1}{c}{\textbf{Metrics}}&\multicolumn{1}{c}{\textbf{Mode}}&\multicolumn{1}{c}{\textbf{WD50K}}&\multicolumn{1}{c}{\textbf{WikiPeople}}&\multicolumn{1}{c}{\textbf{JF17K}} \\
\hline
\multirow{2}{*}{FLOPs} & w/ MoE &118.397G &118.167G &118.070G \\
& w/o MoE &286.851G &286.621G &286.524G \\
\hline
\multirow{2}{*}{params} & w/ MoE &66.956M &66.956M &66.956M \\
& w/o MoE &79.877M &79.877M &79.877M \\
\bottomrule
\end{tabular*}
\label{table_flops_and_params}
\end{table}
\\
\noindent
\zhiweihu{\textbf{Label Smoothing.}}
\vic{The label smoothing strategy has been successfully used in  the KGC task~\citep{Zhiqing_2019,Zhanqiu_2020}. 
It mitigates the bias of the pre-trained data due to random sampling. In Figure~\ref{figure_ablation_studies}(d), we observe that 
setting different label smoothing values  brings subtle performance differences, 
showing that HyperFormer  is robust to the label smoothing strategy, and therefore not causing performance gaps due to improper value selection. In addition, we note that there is no unique label smoothing value that works for all datasets.}
\\ \zhiweihu{\noindent\textbf{Parameters and Computational Complexity.}}
\zhiweihu{The number of parameters can be used to evaluate the trainable parameters of a model, while computational cost refers to the number of floating-point operations (FLOPs) required during training or inference. Table~\ref{table_flops_and_params} shows that introducing the MoE mechanism can simultaneously reduce the number of parameters and computational cost, with a more significant reduction in computational cost. This is intuitively explained by the fact that MoE replaces the feed-forward layers in the original transformer. So, when obtaining the final result, only the predictions from experts with higher confidence are considered, suppressing the involvement of irrelevant neurons in the entire computation process. As a result, both the number of parameters and computational cost are reduced simultaneously.}
\label{section_ablation_study}

\section{Conclusion and future work}

\zhiweihu{In this paper, we proposed  HyperFormer, a framework for the HKGC task which strengthens the bidirectional interaction between entities, relations, and qualifiers, while retaining the structural information of qualifiers in a local-level sequence. Experiments under different conditions on the WD50K, WikiPeople, and JF17K datasets show that HyperFormer achieves  in most cases better results than existing models. The ablation experimental results demonstrate the effectiveness of each module of HyperFormer. For future work, we will integrate other types of data in a KG, e.g., entities's textual descriptions or literals, for better entity representation, and apply HyperFormer into larger scale HKGs like the full WikiData.}

\section*{Acknowledgments}
This work has been supported by the National Natural Science Foundation of China (No.61936012, No.62076155), by  the Key Research and Development Program of Shanxi Province (No.202102020101008),  by the Science and Technology Cooperation and Exchange Special Program of Shanxi Province (No.202204041101016),  by the Chang
Jiang Scholars Program (J2019032) and by   a Leverhulme Trust Research Project Grant (RPG-2021-140).



\bibliographystyle{ACM-Reference-Format}
\bibliography{sample-base}










\end{document}